# Mining the Knowledge of Weights for Optimizing Neural Network Structures

Mengqiao Han, Xiabi Liu, Zhaoyang Hai, SAID, Xin Duan

## Abstract

*Knowledge embedded in the weights of the artificial neural network can be used to improve the network structure, such as in network compression. However, the knowledge is set up by hand, which may not be very accurate, and relevant information may be overlooked. Inspired by how learning works in the mammalian brain, we mine the knowledge contained in the weights of the neural network toward automatic architecture learning in this paper. We introduce a switcher neural network (SNN) that uses as inputs the weights of a task-specific neural network (called TNN for short). By mining the knowledge contained in the weights, the SNN outputs scaling factors for turning off and weighting neurons in the TNN. To optimize the structure and the parameters of TNN simultaneously, the SNN and TNN are learned alternately under the same performance evaluation of TNN using stochastic gradient descent. We test our method on widely used datasets and popular networks in classification applications. In terms of accuracy, we outperform baseline networks and other structure learning methods stably and significantly. At the same time, we compress the baseline networks without introducing any sparse induction mechanism, and our method, in particular, leads to a lower compression rate when dealing with simpler baselines or more difficult tasks. These results demonstrate that our method can produce a more reasonable structure. Code: https://anonymous.4open.science/r/Mining-the-knowladge-of-weights-FD52*

## Introduction

Deep Neural network (DNN), as a flexible function approximator, has demonstrated groundbreaking success in a broad range of applications [1, 2]. The architecture of DNN, however, is typically hand-designed in a process known as architectural engineering, which is usually over-parameterized [3, 4] and easily prone to overfitting [5, 6].

In recent years, there has been a huge interest in automatically finding reasonable network structures that save resources while providing excellent generalization capabilities. The widely used solutions include network compression [7, 8, 9, 10, 11], sparse regularization [12, 13], and neural architecture search (NAS) [14, 15, 16].

Network compression aims to obtain a compact network model from a given backbone network [17]. Most compression methods involve fine-tuning and rely on human experience, which might lead to performance degradation [18]. Sparse regularization adds sparse penalty terms to the loss function to guide the weights toward zero during the optimization process. It may impose unreasonable sparsity or hyperparameters on over-parameterized models [19]. NAS attempts to automate the process of the architecture design by searching among a set of smaller well-known building blocks. While the search methods range from reinforcement learning to gradient-based approaches [16, 20], the search space of possible connectivity patterns is still largely constrained.

The essence of network compression is to investigate the knowledge embedded in the network's weights in order to simplify the network structure. Low weights, for example, indicate unimportant connections [7], connections that move toward zero over time for training are unimportant [21], and so on. However, the knowledge included in existing compression methods is observed by humans, which may not be very accurate, and relevant information may be overlooked. We think that it is better to mine the knowledge embedded in the network weights automatically in order to optimize the network structure. Based on this idea, we introduce a switcher neural network (SNN) that accepts the weight matrix of a task-specific neural network (called TNN for short) as input and outputs scaling factors to turn off some neurons in TNN while strengthening or weakening others (we show the working mechanism of SNN in Fig.1). Based on the introduced SNN, the TNN training consists of two phases.

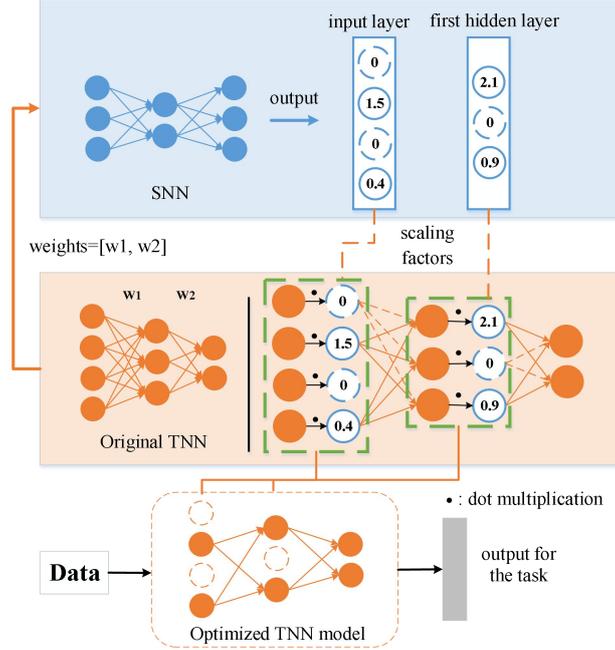

Figure 1. Using SNN to determine the optimal structure of TNN.

In the first phase, the SNN takes the TNN's weights as inputs in order to learn the structural knowledge contained therein, and outputs scaling factors to turn off and weight TNN's neurons to optimize its structure. As a result, neurons are strengthened, weakened, or pruned according to the corresponding scaling factors. In the second phase, the TNN adapts to the new structure by updating its weights using the traditional learning method. SNN and TNN learning phases are iteratively repeated to further optimize network structure and parameters. The loss functions used in these two phases are both the performance evaluation of TNN. Note that we did not introduce any sparse induction, fine-tuning, or pre-training mechanisms. The above training process is similar to how learning works in the mammalian brain. To learn is to myelinate and trigger neuronal changes [22], where myelin and synapses are created in the mammalian brain during early life, followed by gradual pruning of infrequently used connections to typical adult values [23, 24], and myelin insulates the particularly active neuronal axons to increase the speed by which electrical impulses travel along neuronal processes [22].

The main contributions of this paper are summarized as follows:

1. We propose a novel approach for learning neural network (NN) structure by introducing an additional NN (i.e., SNN) in order to mine the knowledge contained in the weights of the TNN. To this end, we design an SNN structure, which is based on the idea of fully exploring the relationship between the weights. And we develop a learning algorithm that alternately optimizes the SNN and TNN under the same learning objective. To our best knowledge, this is the first work that introduces such an approach.

2. We apply the proposed SNN to convolutional neural networks (CNNs) on classification problem and design the corresponding schema for convolutional layers and fully connected (FC) layers, respectively. On widely used public datasets, our approach produces simpler structures and higher accuracies than baselines. And, when compared to other structure learning methods, our approach results in a significant improvement in accuracy.

**Related Work**

In this section, we briefly introduce the main methods used for network structure learning.

**Non-structural parameter pruning.** To compress neural networks, we prune the weak connections in a network, which have a small influence on its prediction. These methods explore the network weights pruning by freezing connections corresponding to small weights [7, 8] or by introducing sparsity regularization [12, 13] to the parameters.

**Structural parameter pruning.** It aims at zeroing out structured groups of the convolutional filters by imposing group sparsity regularization [25, 26, 27, 28]. One branch of works combined the group sparsity regularizer with other regularizer for network pruning [29, 30]. Another branch of research investigated a better group-sparsity regularizer for parameter pruning [31, 32].

**Filter Decomposition.** Tensor decomposition makes the original filter decomposed into a lightweight one and a linear projection, thus resulting in the reduction of parameters and computations [33, 34, 35, 36].

**Neural Architecture Search.** NAS was proposed to explore the space of potential models automatically to optimize the structure in conjunction with the weights [17]. As highlighted by Xie et al. [37], the connectivity patterns in methods of NAS remain largely constrained. Accordingly, Wortsman et al. [38] proposed a method for discovering neural wirings (DNW) – where the weights and structure are jointly optimized free from the typical constraints of NAS. They used the weights with the highest magnitude to choose the subnetwork.

**Other methods.** Knowledge distillation transfers the knowledge of a teacher network to a student network [39]. Current research in this direction focuses on the architectural design of the student network [40] and the loss function [41]. The Bayesian compression leads to prune with sparse inducing prior, and reward uncertain posteriors over parameters through the bits back argument [11, 42]. Furthermore, the key idea of structured dropout is to randomly drop units (along with their connections) from the neural network during training [43, 44].

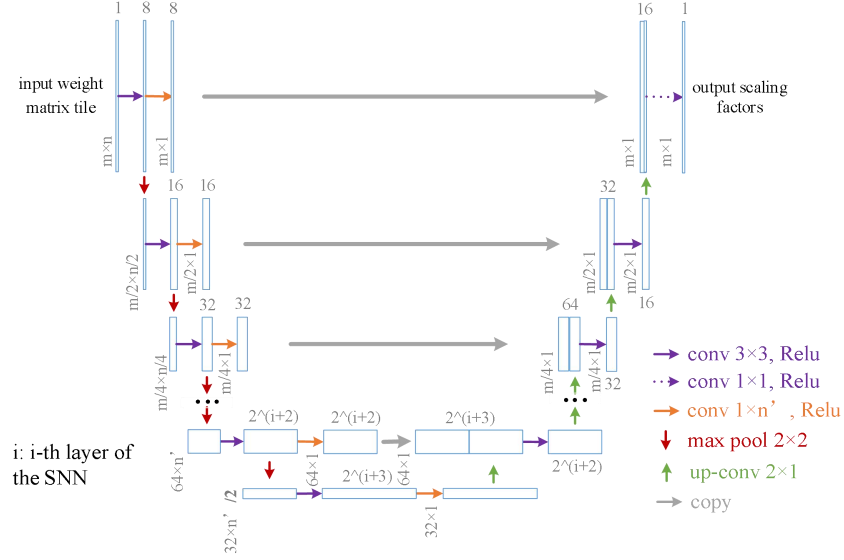

Figure 2. SNN Structure.

## The Proposed Approach

As described above, the core idea of our proposed approach is to mine the knowledge contained in the weights of a NN for determining the importance of each neuron. An SNN is introduced to fulfill this idea, which takes the weights of a TNN as inputs and outputs the scaling factors to turn off and weight the neurons. As a result, some neurons in TNN are pruned, and the others strengthened or weakened.

### 3.1. SNN Structure

The purpose of the SNN is to determine the importance of neurons in the TNN based on the knowledge contained in TNN's weights. The corresponding structure of SNN is designed as in Fig.2, which is explained as follows.

First, we need to extract the features from the weight matrix in order to make decisions to turn off, strengthen, or weaken which neurons. We apply CNN, a powerful tool to extract meaningful features, as the encoder subnetwork which is composed of multiple blocks of convolution, ReLU activation function, and max pooling.

Second, we need to obtain the decision values reflecting the importance of neurons based on the extracted features. We use the decoder to accomplish it. The resultant network is similar to U-Net [45]. Note that the size of SNN outputs is not same as its inputs, since a row of the weight matrix (all connections of a neuron) corresponds to only one scaling factor. Therefore, we need to add a row convolution for each layer of the SNN encoder part to transform the width of the encoded feature to 1 (as in the left of Fig.2) and concatenate it with the corresponding upsampled feature from the deconvolution operator (as in the right of Fig.2).

Finally, the SNN output values through ReLU activation function as scaling factors to turn off and weight neurons.

### 3.2. SNN Guided TNN Optimization

We apply the SNN to guide the structure optimization of the TNN. First, we use SNN to mine the weights in TNN to obtain the structural knowledge contained therein. Then, the outputs of SNN (scaling factors) are multiplied with the outputs of neurons in the TNN.

As a result, the scaling factors play a role of turning off and weighting neurons. There are 3 cases of results:
- "0": pruning, the neurons with zero scaling factors are redundant, we remove all their incoming and outgoing connections and corresponding weights.
- ">=1": strengthening, it means that corresponding neurons are important.
- "<1": weakening, it means that corresponding neurons should be remained, but the degree of importance of it is lower than that of strengthened ones.

Second, the TNN utilizes the structural knowledge mined by the SNN to optimize its weights, in order to adapt to this new structure.

Consequently, the satisfactory TNN is jointly determined by the optimal parameters of SNN and TNN. We present a learning algorithm with two phases to train these two networks together. In the first phase, the TNN parameters are frozen and the SNN parameters are optimized to choose a reasonable structure of TNN. In the second phase, the SNN parameters are frozen and the TNN optimizes its parameters under the guidance of the structural knowledge reflected in the SNN. We use the same loss function in these two phases, such as the cross-entropy (CE) in the classification applications in this paper. By using this strategy, the network sparsity is not required explicitly instead is achieved indirectly by requiring the optimized performance.

The learning method is summarized in Algorithm 1.

---

**Algorithm 1:** The SNN-TNN learning algorithm.

**Initialize:** $W^T$ (weight matrix of TNN), $W^S$ (weight matrix of SNN), T (number of maximal iterations)

**output:** $W^{T*}$ (optimal $W^T$), $W^{S*}$ (optimal $W^S$).

**for** t =0 **to** T do:
   if The loss of SNN-TNN is–still decreasing during the training process:
      if t % 2 == 0:
         Fixing $W^T_t$, $W^S_t$ is updated to $W^S_{t+1}$ through backpropagation.
      else:
         Fixing $W^S_t$, $W^T_t$ is updated to $W^T_{t+1}$ through backpropagation.
   else
      break
**end**

---

### 3.3. Optimizing CNN for classification

We apply our SNN based learning approach for optimizing the structure and parameters of CNN for classification. There are two types of neurons in CNN, convolutional kernels and FC neurons. The corresponding applications of SNN to them are described in detail below.

### 3.3.1 For Fully Connected Neurons

We take a network with two FC layers as an example to illustrate the specific process of introducing SNN into FC layers. Let Wk (k=1,2) denotes the weight matrix of the k-th layer, $\Phi$ denotes the activation function, and X denotes the input data, then the forward computation of this network is

$$f(X,W) = \mathrm{softmax}(\Phi(W_1 X)W_2) \quad (1)$$

All the weights in these two FC layers should be fully considered together and input into the SSN for mining the structural knowledge. This means that we need to concatenate W1 and W2 as a single matrix with two channels. However, the size of W2 is less than that of W1, so we pad W2 with zeros until W2' becomes of the same size as W1. Then the weight matrix [W1, W2'] is taken as the input of SNN. Accordingly, the SNN outputs values to indicate the importance degree of each neuron in these two FC layers.

Finally, each neuron in the input layer and the first hidden layer are multiplied with the corresponding scaling factor, like the way shown in Fig.1.

Let g = {g1, g2} be the SNN outputted scaling factors for the input layer and the first hidden layer, respectively. By introducing our SNN, Eq.1 above is extended to

$$f(X,W) = \mathrm{softmax}(\Phi(X \odot g_1 W_{nl}) \odot g_2 W_{mn}) \quad (2)$$

where $\odot$ denotes the element wise product.

### 3.3.2 For Convolutional Kernels

We explain in detail in this section the application of SNN to the TNN's convolutional layers. The forward calculation for each convolution layer is

$$X_{n+1} = p(\Phi(b_n(W_n X_n))) \quad (3)$$

where $W_n$ denotes the weight matrix of the n-th layer, $b_n$ denotes the Batch Normalization (BN) of the n-th layer, $\Phi$ denotes the activation function, p denotes max pooling, $X_n$ represents the feature map of the n-th layer, $X_{n+1}$ denotes the new feature map created by convolution operation.

We use the SNN to explore the weights in all convolutional layers together. As the number of channels increases in deep CNN, a convolution kernel will have a large number of weights. It is hard to input all of these weights into the SNN. Therefore, we extract the features from the weights of each convolutional layer through adding a convolution calculation and concatenate these features for all the convolutional layers. Then we obtain a single matrix as the input of the SNN, each row of which corresponds to the features of a convolution kernel. This is the only difference between the SNN for the convolutional layers and the SNN for the FC layers.

Similarly, the SNN outputs scaling factors to turn off and weight the neurons of convolutional layers in the TNN. Let $g_n$ be the scaling factors for the n-th convolutional layer, then the output of the convolution layers based on the SNN is changed from Eq. 3 to

$$X_{n+1} = p\left(\Phi\left(b_n\left(W_n X_n\right)\right) \Theta g_n\right) \quad (4)$$

## 3.4. Analysis

We discuss how the network optimizes its weights after the introduction of SNN. To facilitate the discussion, we consider the TNN and the SNN to be the simplest single-layer fully connected networks. This TNN has only one input and K outputs for classification. Correspondingly, the SNN has K inputs and one output for turning off or weighting the input neuron of TNN.

Let D be a dataset consisting of one input-output pairs {(x, y)}, $w^T$ be TNN`s weights, $w^S$ be SNN`s weights, θ be the parameters that consists of TNN`s weights ($w^T$) and SNN`s weights ($w^S$). s and h are used to denote the forward output of SNN and SNN-TNN, respectively, then the gradient of the loss function R(θ) with respect to $w^S$ and $w^T$ are

$$\Delta w_k^S = \frac{\partial R(\theta)}{\partial w_k^S} = w_k^T \sum_{i=0}^{K}\left(\left(h_i(\theta; x) - y_i\right) x w_i^T\right) \quad (5)$$
and
$$\Delta w_k^T = \frac{\partial R(\theta)}{\partial w_k^T} = \left(h_k(\theta; x) - y_k\right) xs + w_k^S \sum_{i=0}^{K}\left(\left(h_i(\theta; x) - y_i\right) x w_i^T\right) \quad (6)$$

In traditional learning without SNN, the gradient of $R(w^T)$ relative to $W^T$ is

$$\Delta w_k^T = \frac{\partial R(w_k^T)}{\partial w_k^T} = \left(h_k(w^T; x) - y_k\right) x \quad (7)$$

By comparing Eq.6 with Eq.7, we can see that without SNN, the weights update is affected by the error related to only one class; whereas with SNN, except this error, the errors for all the classes are considered. Although a weight is only connected to one output neuron, its change has an effect on the classification results. As a result, taking into account the errors of all classes when updating the weights is more reasonable. Furthermore, the errors are weighted by the weights in TNN and SNN. The computation takes into account the relationship between TNN's weights and the importance of the currently updated weight (as reflected by the weight in SNN).

Our method is motivated in part by how learning works in the mammalian brain. The SNN learns the importance of neurons by mining the weights, which has two effects similar to our mammalian brain. First, the pruned neurons are redundant, this is similar to the natural process that gradual prune little-used neurons or connections to typical adult values [22, 23]. Second, the strengthened neurons are analogous to the insulating of membrane. Neuronal cells that are particularly active in the brain, elevated neuronal activity increases oligodendrocyte number in an area of the brain involved in motor learning in vivo [24]. Among them, oligodendrocytes generate myelin, the insulating membrane that covers many neuronal axons and facilitates the propagation of electrical signals along neuronal circuits [46].

| Model | Method | Acc (%) | Pruned architecture | Params (%) |
| --- | --- | --- | --- | --- |

| Model | Method | Acc (%) | Pruned architecture | Params (%) |
|---|---|---|---|---|
| MLP | Baseline [52] | 98.36 | 784-300-100 | - |
| | BC-GNJ [11] | 98.20 | 278-98-13 | 89.24 |
| | BC-GHS [11] | 98.20 | 311-86-14 | 89.45 |
| | L0$_{hc}$ [12] $\lambda$ sep. | 98.20 | 266-88-33 | 90.06 |
| | SNN | **98.41**±0.02 | 456-134-45 | 74.61 |
| LeNet5-Caffe-20-50-800-500 | Baseline [53] | 99.20 | 20-50-800-500 | - |
| | BC-GNJ [11] | 99.00 | 8-13-88-13 | 99.05 |
| | BC-GHS [11] | 99.00 | 5-10-76-16 | 99.36 |
| | L0$_{hc}$ [12] $\lambda$ sep. | 99.00 | 9-18-65-25 | 98.57 |
| | SNN | **99.27**±0.04 | 12-37-268-192 | 84.95 |

Table 1: Results on MNIST dataset, where we show "mean±std" in test accuracy.

| Model | Method | Acc (%) | Pruned architecture | Params (%) |
|---|---|---|---|---|
| VGG-cov2 | Baseline [55] | 78.90 | 64-64-16384-256-256 | - |
| | edge-popup Algorithm [55] | 78.70 | - | 49.20 |
| | SNN | **85.13**±0.09 | 49-43-12530-203-156 | 39.62 |
| VGG16 | Baseline [56] | 92.52 | 512-4096-4096 | - |
| | NestedNet [57], H | 92.40 | 509-4087-4092 | 0.28 (FCLs) |
| | DEN ($\rho = 0.1$) [17] | 92.31 | 289-2992-2574 | 54.58 (FCLs) |
| | SNN | **92.94**±0.32 | 322-3632-3235 | 31.53 (FCLs) |
| ResNet34 | Baseline [17] | 93.52 | 512 | - |
| | N2N [58] | 93.54 | 189 | 63.09 (FCL) |
| | SNN | **93.81**±0.09 | 247 | 51.76 (FCL) |
| MobileNet V1 (×0.25) | Baseline [59] | 86.30 | 1024 | - |
| | Lottery Ticket [8] | 87.90 | 396 | 61.35 (FCL) |
| | DNW (×0.225) [39] | 89.70 | 475 | 53.61 (FCL) |
| | SNN | **91.00**±0.06 | 584 | 42.87 (FCL) |

Table 3: Results on CIFAR10 dataset, where we show "mean±std" in test accuracy.

| Model | Method | Acc (%) | Pruned architecture | Params (%) |
|---|---|---|---|---|
| ResNet34 | Baseline | 52.71 | 512 | - |
| | SNN | **53.50**±0.10 | 376 | 26.56 (FCL) |
| MobileNetV1 (×0.25) | Baseline | 46.30 | 1024 | - |
| | SNN | **48.80**±0.12 | 816 | 20.31 (FCL) |

Table 4: Results on Mini-ImageNet dataset, where we show "mean±std" in test accuracy.–

## 4. Experiments

In this section, we apply the SNN to optimize six representative TNNs: MLP and LetNet5-Caffe-800-500 on MNIST [47], VGG-cov2, VGG16, ResNet34 and MobileNetV1 on CIFAR10 [48], along with ResNet34 and MobileNetV1 on Mini-ImageNet [49].

We carry out the experiments on a GeForce GTX 1080Ti GPU, which has only 12G of memory and cannot support us to optimize all layers in too deep CNNs. So, we optimize each layer including FC layers and convolutional layers for MLP, LetNet5-Caffe-800-500, and VGG_cov2 among the tested networks, while optimize only FC layers for the other networks.

In all of the experiments, we utilize stochastic gradient descent (SGD) as optimizer with a batch size of 64. It starts with a learning rate of 0.1. For the random initialization, we use Kaiming normal distribution [50] for the convolutional layers, and fill the FC layers with values drawn from the normal distribution (mean=0, $std^2$=0.01). The experiments are conducted in the PyTorch platform [51]. We conduct each group of the experiment to show mean and standard deviation (std) of accuracy over 8 runs, as well as the mean of sparsity rates, to evaluate our approach. Note that the experimental results of baseline networks and other network structure learning methods for comparison purposes in the following experiments, are all referred from the corresponding papers.

### 4.1. MNIST CLASSIFICATION

The MNIST dataset contains 60,000 hand-written digits for training and 10,000 for testing, divided into 10 classes. We consider two classical network structures as baselines. MLP has two hidden layers, with 300 and 100 neurons for each. LetNet5-Caffe-800-500 has three convolutional layers and two FC layers with 800 and 500 neurons for each.

#### 4.1.1 Comparisons

In this experiment, we compare the performance of our approach with those of other structure learning approaches reported on the same NNs, including $L_0$ regularization [12] for sparsity learning, and Bayesian compression [11] for network compression.

The obtained learning results are reported in Table 1, including the classification accuracy of trained NNs and the pruned architectures. On the MLP, all the compared structure learning methods can prune to 1/10 of the original network parameters, which is higher than our 1/4 compression ratio. However, compared to the baseline method, the L0 regularization and Bayesian compression showed slightly lower accuracies, while our method brings the improvement of accuracy 0.05%. In the case of LetNet5-Caffe-800-500, the compared methods found out more sparse subnetwork models with the accuracy loss. On the contrary, our method still has a stable improvement of 0.07% compared with the baseline. Although the compression ratio by using our method is lower than other structure learning methods, but it stably and significantly improves the classification accuracies and without introducing any sparsity induction mechanism. This demonstrates that a beneficial network structure should have reasonable sparsity rather than being as sparse as possible.

#### 4.1.2 Visual Analysis

In this section, we conduct an experimental analysis on the feature maps of the first convolutional layer in LetNet5-Caffe-800-500 after SNN optimization in order to show in which parts of the image our learned structure focus on. Fig.3 shows an example for each category of digits in MNIST dataset. We randomly select three strengthened (indicated by red dashed line in Fig.3) and two pruned (indicated by blue dashed line in Fig.3) feature maps for comparison. The feature maps shown in each column come from the same convolution kernel. We can see that the pruned features reflect the global information of the image. Since the global characteristics of categories have been reflected better in some strengthened features, those global features in pruned neurons can be considered being redundant. In addition, local texture information is reflected in other strengthened features.

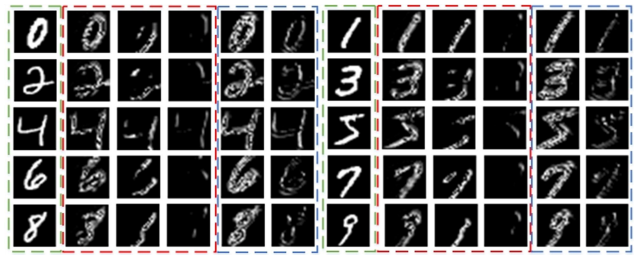

Figure 3. The feature maps are strengthened or pruned in the first convolutional layer of LetNet5-Caffe-800-500.

We further show the scaling pattern produced by SNN for the input neurons of MLP in Fig.4 (left), where the yellow, green, and black regions indicate the cases of strengthening, weakening, and pruning, respectively. The input neurons in MLP are actually corresponding with each pixel in the image, thus Fig.4 (left) shows which part of the image is found important or redundant by using our SNN. We can see from Fig.4 (left) that the strengthened neurons correspond to the center of the image and the pruned neurons exist in the top and bottom edge regions. It is partly in accordance with the fact that digits are written in the center of the image, thus it demonstrates the reasonability of the structure learned by our approach. Why we don't prune the left and right edge regions still need further investigation in the future.

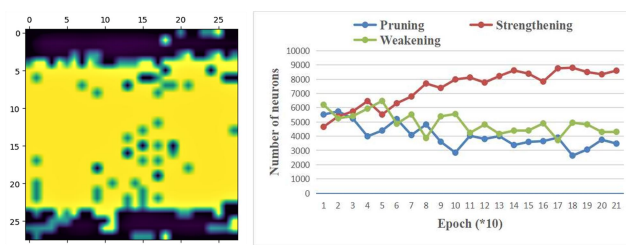

Figure 4. Left: visualize the scaling factors outputted by our SNN for the neurons of the first FC layer in the MLP. Right: the SNN turns off and weights the neurons in the first FC layer of VGG-cov2 during training.

### 4.1.3 Ablation Experiment of Alternate Learning

To verify our alternate learning algorithm, we compare it with another separate learning strategy as follows. First, we train TNN solely and sufficiently using the traditional learning method. Then, we add SNN to improve the structure of this trained TNN. In this stage, only SNN is trained and TNN parameters are frozen. The MLP is used as TNN in this experiment. Table 2 lists the comparison results over 8 runs, where SNN-AL and SNN-SL denote the SNN with alternate learning and the SNN with separate learning, respectively. We can see that the SNN-SL model can still utilize the knowledge of

NN's weights, but it is obviously inferior to the SNN-AL model in terms of accuracy and compression ratio. We think this is reasonable since the production of myelin and neuronal changes are highly dynamic [54].

| Three patterns | Acc (%) | Pruned architecture | Params (%) |
|---|---|---|---|
| MLP [52] | 98.36 | 784-300-100 | - |
| SNN-SL | 98.37±0.03 | 756-287-92 | 8.23 |
| SNN-AL | **98.41**±0.02 | 456-134-45 | 74.61 |

Table 2. Ablation study: our proposed method is applied to MLP and tested on MNIST.

### 4.2. CIFAR10 CLASSIFICATION

CIFAR10 dataset consists of 50,000 training images and 10,000 test images, divided into 10 classes. We validate the accuracy and pruning result of our models on the well-known architectures of VGG-cov2, VGG16, ResNet34 and MobileNetV1.

#### 4.2.1 VGG-cov2

VGG-cov2 has two convolutional layers and three FC layers with 16,384, 256, and 256 neurons in each. The performance of our approach for optimizing this model is given in Table 3. We compare the performance of our approach with the edge-popup algorithm [55] which is a network compression method. It can be seen that the compression ratio of our method (39.62%) is lower than 49.20% of the edge-popup algorithm, but the accuracy can be increased by 6.23% by using our SNN, which is an exciting result.

We analyze why VGG-cov2 can produce such an exciting result. Fig.4 (right) depicts the change in the number of neurons corresponding to three types of SNN outputs, pruning, weakening, and strengthening, along with the epoch iteration. We can see that the number of pruned and weakened neurons is decreased and that of strengthened neurons is increased. The reason behind this phenomenon is that CIFAR10 is a challenging dataset for a relatively simple network like VGG-cov2. Therefore, this network structure shouldn't be compressed too much, and the important neurons should be further strengthened. Fig.4 (right) proves that our SNN can cause neurons in an activation or inhibition state that is more suitable for the data.

#### 4.2.2 Other Networks

The other networks in this experiment, including VGG16, ResNet34, and MobileNetV1, are much complicated than VGG-cov2. They have more than 10 convolution layers. As mentioned above, we use our SNN to optimize only their FC layers because of the limitation in used GPU. Therefore, we only compare the compression ratio on the FC layers with other methods.

We compare our approach with several classic works on structure learning, which were also tested on the same networks, including NestedNet [57], N2N [58], and Lottery Ticket [8] for compressing, as well as DEN [17] and DNW [39] for NAS. The testing results are shown in Table 3. The results show that the classification accuracy on all 3 networks can be improved by using our approach. Compared with N2N for ResNet34, our method has an improvement (0.29%) in accuracy. For MobileNetV1, our SNN surpasses the accuracy of baseline by 4.7%, higher than Lottery Ticket by 1.6%, and DNW by 3.4%. On the VGG16 model, our method achieves better accuracy than the baseline, while NestedNet (H) and DEN methods result in a slight decrease in the same case.

Although in terms of compression ratio, the SNN has a gap of 11%~23% with other structure learning methods, however, our method can achieve more outstanding performance between accuracy and compression ratio. These results prove that the structural knowledge obtained by mining the weights in NN is more effective than that by introducing some artificial settings.

### 4.2.3 Mini-ImageNet CLASSIFICATION

The Mini-ImageNet dataset contains 100 classes randomly chosen from the ImageNet ILSVRC-2012 challenge with 600 images per class. It is split into 64 base classes, 16 validation classes, and 20 novel classes. We apply all the data to the classification tasks, in which 500 images are randomly selected from each category for training and the remaining 100 images for testing. We consider the classical ResNet34 and MobileNetV1 as baseline networks in this application. Same as before, only FC layers are optimized by using our SNN.

Table 4 lists the accuracy and the pruned FC structures achieved by our approach. We can see that the SNN significantly improves the baseline accuracy of ResNet34 and MobileNetV1 by 0.79% and 2.5%, respectively.

When comparing the compression ratios of ResNet34 and MobileNetV1 in Table 3 to those in Table 4, an interesting phenomenon can be observed: our SNN method results in a lower compression ratio of the same network in more difficult tasks. It is further confirmed by Fig. 5, where we visualize the three types of outputs of SNN for ResNet34 and MobileNetV1 in two applications. The yellow, green, and black regions in Fig. 5 correspond to the cases of strengthening, weakening, and pruning, respectively. It is obvious from Fig. 5 that the lower compression ratio of the same network is obtained in more difficult tasks, and more neurons are in an enhanced state. These demonstrate the reasonability of our SNN for learning structures.

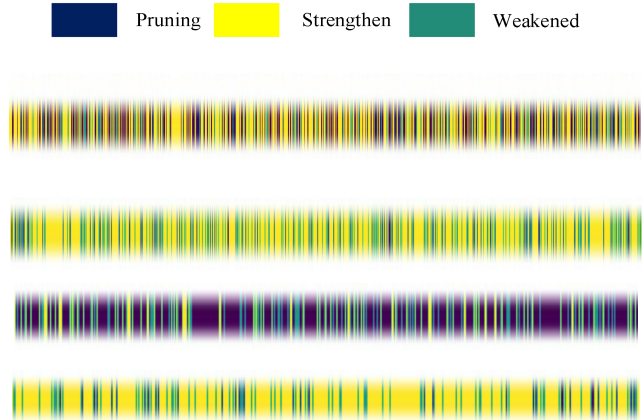

Figure 5. Visualize the output patterns of SNN corresponding to the neurons of the FC layer in MobileNetV1 on CIFAR10 and Mini-ImageNet (the first and the second row), and ResNet34 on CIFAR10 and Mini-ImageNet (the third and the fourth row).

## 5 Conclusions

In this paper, we have proposed a novel approach to optimize the structure of NNs with improving accuracy by mining the knowledge embedded in NN`s weights. Our method, which was inspired in part by how learning works in the mammalian brain, involves turning off and weighting neurons by learning the importance degree of neurons. We tested our proposed approach on popular CNNs on widely used datasets including MNIST, CIFAR10, and

Mini-ImageNet. The obtained results using our SNN show that potential structural knowledge can be mined from network`s weights, which helps us to obtain a more effective and efficient network structure. In the future, we will investigate the possibility of applying SNN to optimize convolutional layers of deep CNNs and try to improve the structure of SNN to get better performance.